%% file: main.tex
\pdfoutput=1

\documentclass[11pt]{article}

\usepackage[final]{acl}
\usepackage{ulem}
\usepackage{enumitem} 
\usepackage[linesnumbered,ruled,vlined]{algorithm2e}
\usepackage{times}
\usepackage{latexsym}
\usepackage{xspace}

\usepackage{amsmath}   
\usepackage{amssymb}   
\usepackage{amsfonts}  
\usepackage{tcolorbox}  

\usepackage[T1]{fontenc}

\usepackage[utf8]{inputenc}

\usepackage{microtype}

\usepackage{inconsolata}

\usepackage{graphicx}
\usepackage{booktabs}
\usepackage{multirow}

\usepackage{xcolor}
\makeatletter
\renewcommand\@makefnmark{%
  \hbox{\textsuperscript{\normalfont\color{black}\@thefnmark}}%
}
\makeatother

\title{\model{}: Rethinking Reinforcement Learning and\\Test-Time Scaling for Efficient and Enhanced Video Reasoning}

\author{%
Ziyang Wang$^{1,}$\thanks{Equal contribution.} \quad 
Jaehong Yoon$^{1, 2, *}$ \quad
Shoubin Yu$^1$ \quad  \\ 
\textbf{Md Mohaiminul Islam}$^1$ \quad  
\textbf{Gedas Bertasius}$^1$ \quad 
\textbf{Mohit Bansal}$^1$ \\ 
\\$^1$UNC Chapel Hill \quad\quad $^2$Nanyang Technological University\\
\\
{{ \tt \normalsize \href{https://sites.google.com/cs.unc.edu/videorts2025/}{\textcolor{magenta}{https://sites.google.com/cs.unc.edu/videorts2025/}} }}}

\input{marcos}

\begin{document}
\maketitle
\input{sec/0_abstract}    
\input{sec/1_intro}

\input{sec/2_related_work}

\input{sec/3_method}

\input{sec/4_exp_setup}

\input{sec/5_results}

\input{sec/6_conclusion}

\section*{Acknowledgments}
We thank the reviewers and area chair, as well as Justin Chen, David Wan, Ce Zhang and Yan-Bo Lin for their helpful discussions. 
This work was supported by DARPA ECOLE Program No. HR00112390060, NSF-AI Engage Institute DRL-2112635, ARO Award W911NF2110220, ONR Grant N00014-23-1-2356, Capital One Research Award, the Accelerate Foundation Models Research program, Laboratory for Analytic Sciences via NC State University, and Sony Focused Research Award. The views contained in this article are those of the authors and not of the funding agency.

\bibliography{custom}

\input{sec/8_appendix}

\end{document}

%% file: marcos.tex
\definecolor{darkgreen}{rgb}{0,0.5,0}
\definecolor{azureblue}{rgb}{0,0.5,1}
\definecolor{darkgreen}{rgb}{1,0,0}
\definecolor{color1}{HTML}{006EB8}
\definecolor{color2}{HTML}{009B55}
\definecolor{color3}{HTML}{924DBF}

\usepackage{pifont}

\usepackage[capitalize]{cleveref}
\crefname{section}{Sec.}{Secs.}
\Crefname{section}{Section}{Sections}
\Crefname{table}{Table}{Tables}
\crefname{table}{Tab.}{Tabs.}
\crefname{appendix}{Sec.}{Secs.}
\Crefname{appendix}{Section}{Sections}

\hypersetup{
    colorlinks=true,
    linkcolor=magenta,
    filecolor=magenta,  
    urlcolor=black,
    citecolor=color3,
    pdftitle={Overleaf Example},
    pdfpagemode=FullScreen,
    }

\usepackage{color,colortbl}
\definecolor{darkgreen}{rgb}{0,0.5,0}

\definecolor{darkgreen}{rgb}{1,0,0}

\definecolor{azureblue}{rgb}{0,0.5,1}

\newcommand{\ours}{\textsc{Video-RTS}\xspace}
\newcommand{\model}{\ours}

%% file: sec/0_abstract.tex
\begin{abstract}
Despite advances in reinforcement learning (RL)-based video reasoning with large language models (LLMs), data collection and fine-tuning remain significant challenges. 
These methods often rely on large-scale supervised fine-tuning (SFT) with extensive video data and long Chain-of-Thought (CoT) annotations, making them costly and hard to scale.
To address this, we present \ours{}, a new approach to improve video reasoning capability with drastically improved data efficiency by combining data-efficient RL with a video-adaptive test-time scaling (TTS) strategy.
Building on observations about the data scaling, we skip the resource-intensive SFT step and employ efficient pure-RL training with output-based rewards, requiring no additional annotations or extensive fine-tuning.
Furthermore, to utilize computational resources more efficiently, we introduce a sparse-to-dense video TTS strategy that improves inference by iteratively adding frames based on output consistency.
We validate our approach on multiple video reasoning benchmarks, showing that \model surpasses existing video reasoning models by $2.4\%$ in accuracy using only $3.6\%$ training samples. 
Specifically, \model{} achieves a $4.2\%$ improvement on Video-Holmes, a recent and challenging video reasoning benchmark.
Notably, our pure RL training and adaptive video TTS offer complementary strengths, enabling \model{}'s strong reasoning performance.
\end{abstract}

%% file: sec/1_intro.tex
\section{Introduction}
\label{sec:intro}

\begin{figure*}
    \centering
    \includegraphics[width=1\linewidth]{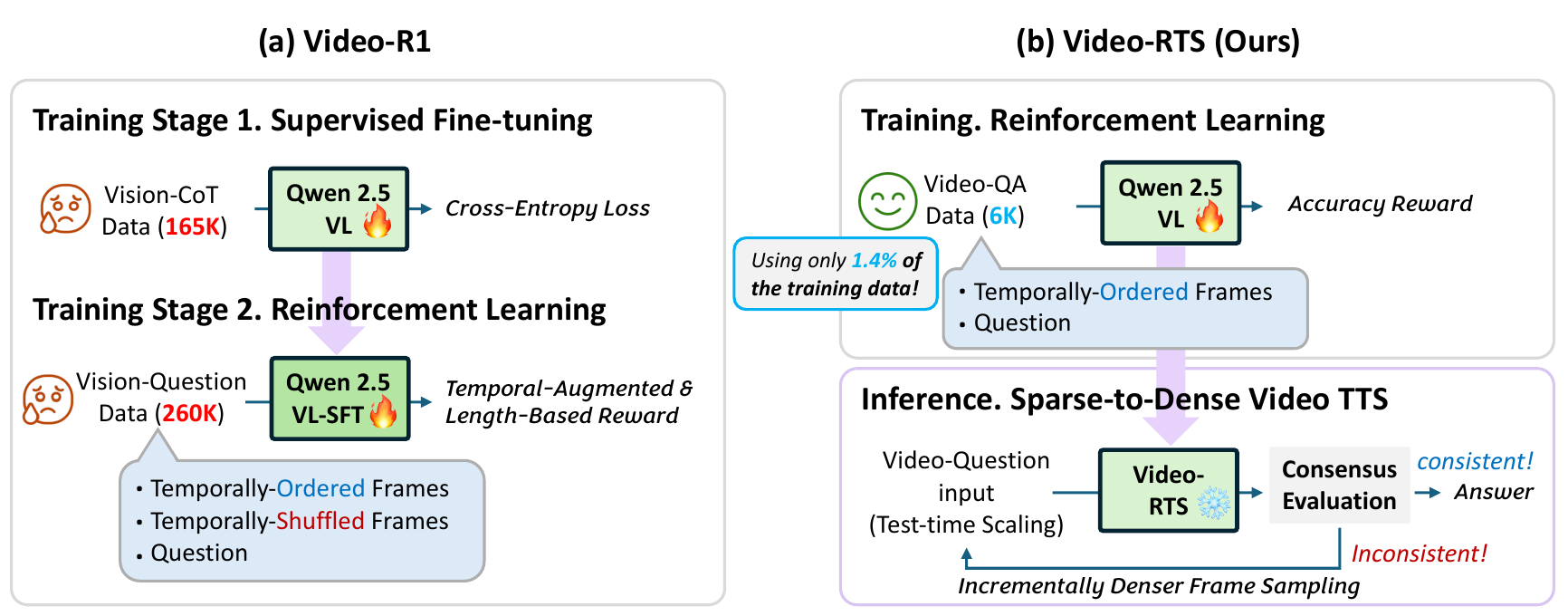}
    \caption{\textbf{Training and inference recipe comparison between Video-R1~\cite{feng2025video} and our \ours{}.}
    While (a) Video-R1 uses a two-stage pipeline with SFT and RL, (b) \ours{} adapts a pure-RL approach with output-based rewards for better data efficiency. We further enhance the reasoning of \ours{} by proposing dynamic sparse-to-dense video test-time scaling. The format reward is omitted, as both models use it.
    } 
    \label{fig:intro-comparison}
\end{figure*}

Large language models (LLMs) have demonstrated strong problem-solving abilities across diverse domains, enabled by techniques such as Chain-of-Thought (CoT) reasoning~\citep{wang2022self, yao2023tree} and multi-agent collaboration \cite{talebirad2023multiagentcollaborationharnessingpower, chen2024reconcileroundtableconferenceimproves}.
Building on advances in the language domain, several approaches~\citep{liu2025videomind, fei2024videoofthoughtstepbystepvideoreasoning, feng2025video,sun2025video,li2025videochat} have recently extended them to improve video reasoning capabilities. 
However, these methods demand high computational costs and lower training efficiency, typically following an extensive two-stage recipe: (\emph{i}) supervised fine-tuning (SFT) on reasoning-focused prompts with step-by-step chain-of-thought annotations, followed by (\emph{ii}) large-scale reinforcement learning using rewards over massive collections of video question-answering data. 
This pipeline poses substantial computational overhead, particularly in generating long CoT data for video corpus, which limits its scalability for complex, long-term video reasoning tasks.

To overcome these limitations and enable efficient video reasoning, we propose \model{}, a novel approach that integrates data-efficient reinforcement learning with video-adaptive test-time scaling strategies, significantly enhancing reasoning performance while maintaining efficiency. 
In training, unlike the existing approaches \cite{wang2025videorftincentivizingvideoreasoning, feng2025video}, which rely on large-scale supervised fine-tuning (SFT) data with long CoT annotation, we skip the data generation step and directly utilize pure RL training on simple video question-answering (QA) data. 
Specifically, we adapt the outcome-supervised RL (group relative preference optimization, GRPO \citep{shao2024deepseekmath}), motivated by DeepSeek-R1-Zero~\citep{deepseekai2025deepseekr1incentivizingreasoningcapability}, for its simplicity and effectiveness in aligning model outputs with answer correctness. 
With merely $6\text{K}$ video-question pairs for RL, our approach matches the performance of the existing SFT+RL framework (Video-R1 \citep{feng2025video}), which relies on $165\text{K}$ SFT examples plus $4\text{K}$ RL examples, underscoring the effectiveness and training efficiency of \model{}.

Furthermore, as illustrated in \cref{fig:training_sample_fig}, scaling to even more video QA samples only brings marginal improvements, suggesting that the RL training saturates quickly on video reasoning data. 
This matches the recent findings in the language domain \cite{wang2025reinforcement} that very few RL training samples could bring great improvement on reasoning tasks. 
Thus, inspired by the test-time scaling works \cite{wang2022self, yao2023tree, snell2024scaling} in the language community, we aim to enhance the video reasoning capability at the inference stage to better allocate the computational resources. 
To the best of our knowledge, this is the first study to systematically explore the combination of reinforcement learning and test-time inference strategies for improving video reasoning capability.

To better allocate the excessive training computation, we propose a sparse-to-dense test-time scaling mechanism specifically designed for video reasoning. 
Specifically, \model{} adaptively selects the appropriate temporal context based on output consistency by iteratively adding more frames at the inference stage. 
Taking advantage of the pure-RL training, the model is able to generate a diverse deep reasoning process given the challenging video query, which allows us to utilize a self-consistency check to decide whether the model obtains sufficient temporal context.
The combination of efficient training and adaptive inference enables the model to adapt its computational effort based on the complexity of each input query, producing accurate responses while using only the necessary amount of resources.

We evaluate \model{} on the five popular video reasoning benchmarks, including Video-Holmes~\cite{cheng2025video}, Video-MMMU \cite{hu2025videommmuevaluatingknowledgeacquisition}, MMVU \cite{zhao2025mmvumeasuringexpertlevelmultidiscipline}, VideoMME \cite{fu2024video} and LongVideoBench \cite{wu2024longvideobench}. 
Results show that across all benchmarks, compared to the recent Video-R1 model~\cite{feng2025video}, which trained on $169K$ samples, \model{}, trained with only $6K$ samples (i.e., $96.4\%$ fewer samples),
outperforms by $2.4\%$ in average accuracy while using fewer frames during inference.
Specifically, on Video-Holmes, the recently proposed complex video reasoning benchmark, \model{} outperforms Video-R1 by $4.2\%$, demonstrating the efficiency and effectiveness of our framework.
Furthermore, we find that our pure RL training and sparse-to-dense video test-time scaling are complementary: RL enhances the MLLM’s reasoning capabilities, while \model{} leverages diverse reasoning strategies to adaptively select the optimal temporal context (i.e., number of frames) for each video query.

%% file: sec/2_related_work.tex
\section{Related Works}
\label{sec:related_work}

\paragraph{Long Video Understanding.}

The rise of video understanding models has expanded from short videos to long-video tasks such as classification~\cite{wu2021towards, mohaiminul2022long, islam2023efficient}, captioning~\cite{zhou2018towards, krishna2017dense, islam2024video}, and question answering~\cite{fu2024video, zhou2024mlvu, wu2024longvideobench}. The emergence of multimodal large language models (MLLMs)~\cite{Qwen2.5-VL,zhang2024video, li2024llava, wang2024internvideo2, bai2025qwen2, zhang2023simple, islam2025bimba, wei2025premindmultiagentvideounderstanding} has further propelled research in long-video understanding. However, most existing MLLMs focus solely on generating answers without providing rationale or reasoning. We address this gap by proposing a new approach that enables MLLMs to generate both answers and step-by-step reasoning through data-efficient pure-RL training and a video-adaptive test-time scaling mechanism, enhancing interpretability and reducing overfitting.

\paragraph{Visual CoT Reasoning with RL.} 
Inspired by the reasoning capabilities demonstrated by large language models (LLMs) in NLP~\cite{deepseekai2025deepseekr1incentivizingreasoningcapability, openai2024openaio1card}, recent efforts have focused on enhancing the reasoning abilities of MLLMs in visual data. 
Early works targeted image-based reasoning, often using hand-crafted CoT structures \cite{xu2024llava, thawakar2025llamav} and modality bridging techniques~\cite{yang2025r1, huang2025vision}.
On the other hand, in the video domain, some approaches focused on temporal grounding ~\cite{wang2024timerefine}, while others employed manual reasoning pipelines for general video understanding~\cite{liu2025videomind, fei2024videoofthoughtstepbystepvideoreasoning}.
Lastly, several recent works \cite{meng2025videocapr1enhancingmllmsvideo, wang2025timezero, sun2025video, zhang2025tinyllavavideor1smallerlmmsvideo, dang2025reinforcingvideoreasoningfocused, li2025veripo} have employed Reinforcement Learning (RL)~\cite{kaelbling1996reinforcement} strategies such as DPO\cite{DPO} and GRPO~\cite{shao2024deepseekmath} for enhancing MLLM reasoning capabilities. 
However, the leading methods \cite{ wang2025videorftincentivizingvideoreasoning, feng2025video,tian2025egor1chainoftoolthoughtultralongegocentric} often rely on a costly SFT stage with large amounts of long CoT data. 
Instead, we use a GRPO-based RL strategy without requiring any SFT data or expensive temporal ordering supervision, enabling data- and compute-efficient training of the video reasoning model.

\paragraph{Test-Time Scaling (TTS).} 
TTS \cite{snell2024scaling} refers to the strategic allocation of increased computational resources during inference to facilitate more deliberate, step-by-step reasoning rather than rapid, heuristic processing. 
While the Chain-of-Thoughts (CoT) framework initially proposed augmenting computational budget during inference to enhance reasoning capabilities, more sophisticated methods have since emerged for language tasks, including self-consistency~\cite{wang2022self}, weighted voting~\cite{wan2025reasoningawareselfconsistencyleveraging}, Tree-of-Thoughts \cite{yao2023tree} and self-reflection~\cite{shinn2023reflexionlanguageagentsverbal}. 
However, we argue that text-centric approaches are sub-optimal on complex video understanding tasks, as they overlook the unique characteristics of videos and the varying levels of reasoning granularity required by different queries. 
Thus, we propose a novel video-adaptive test-time scaling mechanism tailored for the challenges of efficient long-range video reasoning task. 
Specifically, \model{} dynamically allocate inference budget by a sparse-to-dense manner based on the model output consistency.

%% file: sec/3_method.tex
\begin{figure}
    \centering
    \hspace*{-0.4cm}
    {\includegraphics[width=1.05\linewidth]{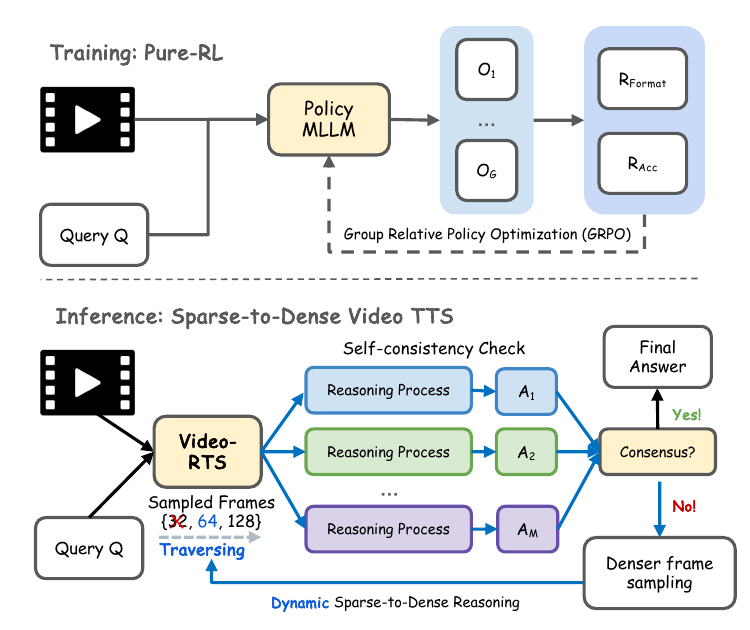}}
    \caption{\textbf{The overview of \ours{}.} The training phase (Top) adapts GRPO-based RL to optimize the MLLM with outcome and accuracy rewards. In inference (Bottom), \ours{} conducts dynamic sparse-to-dense reasoning by traversing sampled frames for generating rationales. If answers are in consensus, it returns an answer; otherwise, it samples denser frames.}
    \label{fig:method}
\end{figure}

\section{\ours{}}
\label{sec:method}
We propose Video-RTS, a resource-efficient RL and test-time scaling framework for video reasoning. 
We begin by introducing outcome-supervised RL method, which serves as our base RL algorithm~\cref{subsec:GRPO}.
In~\cref{subsec:challenges}, we define the problem statement and challenges of the video reasoning problem. 
Next, we propose an efficient reinforcement fine-tuning strategy that leverages simple video QA data without costly chain-of-thought annotations or temporal labels in \cref{subsec:rl}. Finally, we introduce a video-specific test-time scaling mechanism that adaptively adjusts computation, further enhancing performance in \cref{subsec:tts}.

\subsection{Preliminary: Group Relative Policy Optimization (GRPO)}
\label{subsec:GRPO}
Recently, DeepSeek-R1~\citep{deepseekai2025deepseekr1incentivizingreasoningcapability} achieves state-of-the-art performance on multiple language reasoning benchmarks with a newly suggested reinforcement learning (RL) approach.
As a key step of the framework, DeepSeek-R1 utilizes Group Relative Policy Optimization (GRPO)~\citep{shao2024deepseekmath} as the core algorithm to conduct reasoning-oriented RL.
Compared to the canonical DPO~\citep{rafailov2024directpreferenceoptimizationlanguage}, GRPO eliminates the need for a value model by estimating baselines from group-level scores. Directly comparing groups of candidate responses removes reliance on a critic model and substantially reduces training costs.
Given the input question, GRPO first generates $G$ distinct candidate responses $\{O_1, \dots,O_G\}$ through different sampling settings from the old policy $\pi_{\theta_{old}}$.
The model serves as the reward function to get the corresponding scores $\{R_1, \dots, R_G\}$.
Then the model computes the mean and standard deviation of the candidate's score for normalization and determines the quality of these responses:
\begin{equation}
\label{eq:ro}
    S_i=
    \frac{R_i-\mathrm{mean}(\{R_i\}_{i=1}^N)}{\mathrm{std}(\{R_i\}_{i=1}^N)} \text{,}
\end{equation}
where $S_i$ represents the relative quality score of the $i$-th answer candidates.
Given the reasoning question $q\sim P(Q)$, GRPO optimizes the policy model $\pi_{\theta}$ by maximizing the following objective:
\begingroup          
\small     
\begin{align}
\mathcal{J}_{\text{GRPO}}(\theta) = 
&\ \mathbb{E}_{q \sim P(Q), \{o_i\}_{i=1}^{G} \sim \pi_{\theta_{\text{old}}}(O \mid q)}  \notag \\
&\quad \frac{1}{G} \sum_{i=1}^{G} \Bigg[ 
    \min \Bigg( 
        \frac{\pi_{\theta}(o_i \mid q)}{\pi_{\theta_{\text{old}}}(o_i \mid q)} S_i, \notag \\
&\qquad \text{clip} \left( 
        \frac{\pi_{\theta}(o_i \mid q)}{\pi_{\theta_{\text{old}}}(o_i \mid q)}, 
        1 - \epsilon, 1 + \epsilon 
    \right) S_i 
    \Bigg) \notag \\
&\quad\quad - \beta\, \mathbb{D}_{\text{KL}} \Big( \pi_{\theta} \Big\| \pi_{\text{ref}} \Big)\Bigg]. \label{eq:grpo}
\end{align}
\endgroup
To prevent the updated policy $\pi_\theta$, parameterized by $\theta$, from drifting too far from the reference model $\pi_\mathrm{ref}$, GRPO incorporates a KL-divergence term $\mathbb{D}_{\text{KL}}$ that penalizes per-token deviations. In this work, we adopt GRPO as our reinforcement learning algorithm to efficiently enhance video reasoning capabilities.

\subsection{Problem Statement and Challenges}\label{subsec:challenges}
We formulate the video reasoning task as a video question-answering problem, where given a video input $V$ and a reasoning question $Q$, the video reasoning model $f_\theta$ is designed to generate its predicted answer $\widehat{A}$. 
Following the standard practice in the recent video reasoning benchmarks \cite{hu2025videommmuevaluatingknowledgeacquisition, fu2024videommefirstevercomprehensiveevaluation}, we focus on the multiple-choice question-answering format (MCQA), which also adds answer options $A_o$ as input and requires the model $f_\theta$ to choose between the given answer candidates. 
Concretely, the video reasoning process could be formulated as: 
\begin{equation}
    \widehat{A} = f_\theta\bigl(V, Q, A_o\bigr).
\end{equation}
Recently, a few notable works~\citep{feng2025video,sun2025video} show the strong potential of combining supervised fine-tuning (SFT) and reinforcement learning (RL) for addressing video reasoning problems. 
These methods typically follow a two-stage pipeline: (1) SFT with long Chain-of-Thought (CoT) video QA data, and (2) reasoning-focused RL on video QA data. Despite their effectiveness, several challenges remain: (i) \textbf{data inefficiency}: reliance on large-scale video-question or CoT datasets hinders scalability to complex video tasks (e.g., Video-R1 utilize $165K$ SFT data and $24K$ RL data), (ii) \textbf{computational inefficiency during RL}: training with dense video-text pairs and complex reward designs is resource-intensive (e.g., temporal GRPO~\cite{feng2025video}), (iii) \textbf{limited inference-time adaptability}: current models lack mechanisms to scale computation dynamically based on query complexity.

To address these challenges, we develop \ours{}, a data-efficient, yet strong video reasoning model that introduces an advanced training recipe along with a consensus-based hierarchical voting strategy for inference.

\subsection{Resource-Efficient RL for Video Reasoning}\label{subsec:rl}
We introduce the proposed RL training strategy of \ours that overcomes the limitations of machine-generated CoT data and the overhead of supervised fine-tuning.
The pioneering video reasoning approach, Video R1~\cite{feng2025video}, leverages an open-source MLLM (Qwen-2.5-VL-72B \cite{Qwen2.5-VL}) to generate reasoning chains over $165K$ video QA examples for supervised fine-tuning. 
Generating large-scale, long-form reasoning chains is time-consuming, and the quality of the resulting SFT data remains uncertain, as the MLLM shows a significant performance gap compared to human experts and lacks fine-tuning on video-specific CoT reasoning formats.

Motivated by the success of DeepSeek-R1-Zero~\citep{deepseekai2025deepseekr1incentivizingreasoningcapability}, we revisit the standard training pipeline and propose to bypass the costly SFT stage, instead exploring a pure reinforcement learning approach for video QA with minimal training overhead.
To equip the video reasoning capabilities in recent powerful image reasoning MLLMs, we apply outcome-supervised RL (i.e., GRPO) directly on video QA data, using a simple reward function based solely on answer correctness, without relying on any additional verifier.
The details of each component are described below.

\paragraph{Backbone MLLM.} 
As demonstrated in DeepSeek-R1~\citep{deepseekai2025deepseekr1incentivizingreasoningcapability}, an important prerequisite for effective outcome-supervised RL training is the cold-start supervised fine-tuning (SFT) stage, which enhances the model's basic reasoning ability.
Prior works on frame-based video understanding~\citep{buch2022revisiting, lei2022revealingsingleframebias} have shown that models extensively trained on image data can achieve strong performance on video tasks. 
Based on this insight, we use an MLLM (e.g., Qwen-2.5-VL \cite{Qwen2.5-VL}) trained on image reasoning data as a strong cold-started model for outcome-supervised RL training on video data.

\paragraph{Reward Design.} 
Inspired by DeepSeek-R1-zero~\citep{deepseekai2025deepseekr1incentivizingreasoningcapability}, we propose that directly optimizing for outcome-based rewards, rather than relying on step-by-step supervision as in video-SALMONN-o1~\cite{sun2025video}, can further improve reasoning capabilities while reducing the need for costly intermediate CoT data.
Moreover, acquiring detailed supervision for intermediate reasoning steps often demands complex verifier designs.
To address this, we adopt an efficient reward design that directly optimizes the model’s final output $O$.
Specifically, we introduce two types of rewards: \textit{format reward} and \textit{accuracy reward}, to fine-tune the backbone MLLM and induce explicit CoT reasoning ability on complex video tasks.
\begin{itemize}[leftmargin=10pt]
    \item First, we apply a format reward $R_{\mathrm{format}}$ that encourages the model to generate its reasoning process between `<think>' and `</think>' tags before generating the answer prediction. 
    This reward helps the model to have an explicit logical reasoning step in response to the video and text query before producing the final answer.
    \item Next, we introduce an accuracy reward $R_{\mathrm{acc}}$,  which incentivizes the model to produce correct answers following its reasoning process.
    We formulate the training task as a multiple-choice QA problem, enabling a straightforward definition of the reward by comparing the model’s predicted answer $\widehat{A}$ with the ground truth $A$. 
\end{itemize}

\noindent The overall reward function is defined as:
\begin{equation}
\label{eq:rewards}
    R(O) = R_{\mathrm{format}}(O) + R_{\mathrm{acc}}(\widehat{A};A).
\end{equation}

\paragraph{RL Training.} 
As mentioned in \cref{subsec:GRPO}, we adopt the Group Relative Policy Optimization (GRPO)~\cite{shao2024deepseekmath} algorithm for RL on video QA tasks, using the proposed reward functions outlined in \cref{eq:rewards}.
Given an input video $V$ and query $Q$, the model first generates $G$ diverse candidate responses $\{O_1, \dots,O_G\}$ with varied sampling configurations.
Format and accuracy rewards (\cref{eq:rewards}) are then applied to each candidate response to compute their corresponding reward scores $\{R_1, \dots, R_N\}$.
Subsequently, the model (i.e., policy) $\pi_\theta$ is optimized using the GRPO objective as detailed in \cref{eq:grpo}. 
This approach enables efficient RL training for video reasoning using only readily available video-question-answer triplets, with outcome-based rewards that are simple and fast to compute.
Empirically, we find that our RL training design achieves comparable video reasoning performance using just $6K$ samples, compared to existing methods trained on large-scale SFT and RL datasets ($165K$ + $4K$), demonstrating the data and computational efficiency of \model{}.

\subsection{Dynamic Sparse-to-Dense Video Test-Time Scaling}\label{subsec:tts}
With the RL training on video reasoning data, our model is able to generate a long chain-of-thought reasoning process to solve the video reasoning problem. 
However, as shown in \cref{fig:training_sample_fig}, we find that after $6K$ training samples, adding many more video QA samples brings marginal improvements to the video reasoning performance. 
Inspired by the recent progress in test-time scaling technique \cite{snell2024scaling, wang2022self, yao2023tree} from the language community, we aim to save the excessive computational resources in the training stage and allocate them during the inference stage to improve the video reasoning capability. 
Given the redundant nature of video data \cite{wang2024videotree}, an adaptive inference strategy with sparse-to-dense exploration can be both efficient and effective.

To this end, we propose a sparse-to-dense video test-time scaling strategy that iteratively refines the video reasoning process by scaling the frame inputs.  
Inspired by the majority voting method \cite{wang2022self} in the NLP domain, we utilize the self-consistency of the MLLM as the signal of whether the model requires denser information for accurate video reasoning. 
Specifically, given the input video with $n$ frames $V(n)$ and query $Q$, the RL-trained model $\pi_\theta$ generates $m$ different responses $\{O_1, \dots,O_m\}$ given $m$ different sampling parameter for the MLLM.
These responses include diverse reasoning processes on the video input and given query, which provide logical thinking from different angles under the current frame rate. 
Then, we extract the predicted answer $\{\widehat{A}_1, \dots,\widehat{A}_m\}$ from each output and check whether different reasoning process leads to a unanimous answer prediction.  
If the diverse reasoning processes make a consensus, we consider the current temporal information sufficient, and we trust the current prediction. 
If the model generates conflicting predictions, we consider that the current temporal information is not enough for the model to generate an accurate response on the given video and query. 
Thus, we increase the frame rate and conduct the majority voting process iteratively until the model finds a consensus or it reaches the frame rate limit.
We show the detailed algorithm of sparse-to-dense video test-time scaling in \cref{alg:tts}.
With the sparse-to-dense exploration during the inference stage, \model{} adaptively allocates the computational budget for the sample with different temporal requirements and improves the video reasoning performance.

\begin{algorithm}[t]
\small
\caption{Sparse-to-Dense Video TTS}
\label{alg:tts}
\SetKwInOut{Input}{Input}\SetKwInOut{Output}{Output}

\Input{
  Video $\mathbf{V}$ ($N$ frames), query $Q$; policy $\pi_{\theta}$;\\
  sample count $m$; initial frame budget $n_{\text{init}}$;\\
  maximum budget $n_{\max}$
}
\Output{Predicted answer $\hat{A}$}

$n \leftarrow n_{\text{init}}$\;

\While{$n \le n_{\max}$}{
  $\mathcal{V}\bigl(n\bigr) \leftarrow$ first $n$ frames of $\mathbf{V}$\;  
  \For{$i \leftarrow 1$ \KwTo $m$}{
      $O_i \leftarrow \pi_{\theta}\!\bigl(\mathcal{V}(n), Q\bigr)$\;
      $\hat{A}_i \leftarrow \texttt{ExtractAnswer}(O_i)$\;
  }
  \If{$\hat{A}_1 = \hat{A}_2 = \dots = \hat{A}_m$}{
      \Return $\hat{A} \leftarrow \hat{A}_1$\;
  }
  \ElseIf{$n = n_{\max}$}{
      \Return $\hat{A} \leftarrow \texttt{MajorityVote}\!\bigl(\{\hat{A}_i\}_{i=1}^m\bigr)$\;
  }
  $n \leftarrow \min\!\bigl(n * 2,\, n_{\max}\bigr)$\;
}
\end{algorithm}

%% file: sec/4_exp_setup.tex
\section{Experimental Setup}
\label{sec:exp_setup}

\begin{table*}[t]
  \centering
  \setlength{\tabcolsep}{3pt}  
\resizebox{1.9\columnwidth}{!}{ 
  \begin{tabular}{l|c|ccccc}
    \toprule
    \textbf{Method} &
    \textbf{\#Frame} &
    \textbf{Video-Holmes} &
    \textbf{MMVU(mc)} &
    \textbf{Video-MMMU} &
    \textbf{LVB(val)} &
    \textbf{Video-MME} \\
    \midrule
\rowcolor{gray!20} \multicolumn{7}{c}{\textit{Proprietary MLLMs}}\\
    GPT-4o & -& 42.0  & 75.4 &  61.2 &  66.7 & 71.9 \\
Gemini 1.5 Pro & -&  41.3 & 71.2 &  53.4 &  64.0 & 75.0 \\
    \midrule
\rowcolor{gray!20} \multicolumn{7}{c}{\textit{Open-Source General-Purpose MLLMs}} \\

    LLaVA-OV-7B & $64$ & - & 49.2 & 33.8  & 56.3 & 58.2  \\ 
    ViLA-1.5-8B & $64$ & - & 49.2 & 33.8  & 56.3 & 58.2  \\ 

    Qwen-2.5-VL-7B & $\leq 768$ & 27.8 & 59.2 & 47.4  &  56.0 & \textbf{65.1}  \\ 

        \midrule 
\rowcolor{gray!20} \multicolumn{7}{c}{\textit{Video Reasoning LLMs}} \\

    VideoMind-7B & $>64$ & – & – & –  & – & 58.2  \\ 
    VideoTree  & $64$ & – & 54.2 & 47.8  & 52.3 & 56.1  \\  
    Video-R1-7B & $64$ & 36.5 & 63.8 & 52.4 & 53.4 & 61.4 \\
    \midrule

\textbf{\model{}-7B (ours)} & \textbf{51.2} & \textbf{40.7} & \textbf{66.4}& \textbf{52.7} & \textbf{56.6} & 63.0 \\

    \bottomrule
  \end{tabular}
  }
  \vspace{-0.05in}
  \caption{Comparison of the overall accuracy (\%) with the state-of-the-art methods on five video reasoning tasks. We \textbf{highlight} the best performance model on 7B scale for each benchmark.}
  \label{tab:videoqa_benchmark_comparison}
  \vspace{-0.05in}
\end{table*}

\subsection{Evaluation Benchmarks}
(1) \textbf{Video-Holmes}~\citep{cheng2025video} is a newly released and challenging benchmark designed to evaluate the complex video reasoning capabilities of MLLMs. 
It consists of $1837$ questions sourced from $270$ manually annotated suspense short films (ranging from 1 to 5 minutes), which span seven carefully curated tasks.

\noindent(2) \textbf{MMVU}~\citep{zhao2025mmvumeasuringexpertlevelmultidiscipline} is a comprehensive expert-level, multi-discipline benchmark for evaluating video understanding. 
We test on the val split of MMVU on multiple-choice QA format, which contains $625$ QA samples that require expert-level reasoning on complex videos. 

\noindent(3) \textbf{VideoMMMU}~\citep{hu2025videommmuevaluatingknowledgeacquisition} is a multi-modal and multi-disciplinary video benchmark that evaluates LMMs' knowledge acquisition capability from videos. 
We use the standard split, which contains $900$ video reasoning questions in perception, comprehension, and adaptation tasks. 

\noindent(4) \textbf{Video-MME}~\citep{fu2024videommefirstevercomprehensiveevaluation} is a recently proposed comprehensive evaluation benchmark for video analysis from short to long videos ((avg. 17 min)). 
We use the standard split of Video-MME, which contains $2700$ expert-labeled QA pairs designed for both perception and reasoning tasks.

\noindent(5) \textbf{LongVideoBench (LVB)}~\citep{wu2024longvideobench} is a video QA benchmark that highlights referred reasoning questions, which are dependent on long frame inputs. 
We test on the public validation split, which contains $1337$ video reasoning questions.

\subsection{Evaluation Metrics}
We evaluate \model{} on all datasets under the multiple-choice QA setting. We utilize standard accuracy metrics for all experiments. 

\subsection{Training Data}

We leverage CG-Bench \cite{chen2024cgbenchcluegroundedquestionanswering} as training data, which originally contains $12K$ MCQA data. 
Following \citet{yu2025dapoopensourcellmreinforcement, zheng2025deepeyesincentivizingthinkingimages}, we filter out the samples that are too easy or too hard for effective learning. 
Specifically, we generate 8 responses per sample and calculate the difficulty based on the accuracy, samples with accuracy of either 0 or 1 are excluded.
We finally sample a subset of $6K$ MCQA pair for training. 

\subsection{Implementation Details}
We use Qwen-2.5-VL-7B~\citep{Qwen2.5-VL} as our base MLLM. 
To speed up the training process, we uniformly sample $32$ frames for each video and resize the short side of the video to $224$ resolution while keeping the original aspect ratio. 
For GRPO-related implementation, we reference the TRL \cite{vonwerra2022trl} library. 
We train our model with $1$ epoch and only fine-tune the LLM's parameters. 
For $8$ NVIDIA-H100 GPUs, the training takes approximately half a day to finish. 
For hyperparameters, we follow the recent works \cite{zhang2025r1} and set the learning rate as $1e-6$ and the batch size as $16$, the $\beta$ for KL is set to $0.04$. 
For all voting methods, we set the sample number $m$ as $5$. 
For evaluation, we set the base frame number as $32$ as the default. 
For knowledge-focused benchmarks (MMVU, Video-MMMU), 
We set the max frame number as $64$ for knowledge-focused benchmarks (MMVU, Video-MMMU) and $128$ for general long video (reasoning) benchmarks (Video-Holmes, Video-MME, LVB).

%% file: sec/5_results.tex
\section{Experimental Results}
\label{sec:results}

\subsection{Comparison with State-of-the-Art}

\cref{tab:videoqa_benchmark_comparison} shows a comparison of the existing works and \model{} on five popular video reasoning benchmarks. We compare our methods with three types of models: leading proprietary MLLMs \cite{openai2024gpt4ocard, geminiteam2024gemini15unlockingmultimodal}, open-source general-purpose MLLMs \cite{li2024llava, Qwen2.5-VL} and video reasoning LLMs \cite{liu2025videomind, wang2024videotree, feng2025video}. 
Specifically, our approach achieves an accuracy of $40.7\%$ on the Video-Holmes benchmark, outperforming the best open-source 7B model by a significant margin of $4.2\%$, and performing comparably to proprietary models such as Gemini 1.5 Pro and GPT-4o. 
These results validate the effectiveness of \model{} towards complex video reasoning tasks.
On the benchmarks that require expert-level reasoning ability over the complex videos (MMVU, Video-MMMU), \model{} outperforms the SOTA video reasoning model trained on $169K$ total training samples (Video-R1 \cite{feng2025video}) by $2.6\%$ and $0.3\%$ using only $6k$ samples. 
Meanwhile, our methods only utilize $42.8$ and $45.2$ average frames for inference, validating the frame efficiency of \model{} that provides by the adaptive video test-time scaling strategy at inference stage. 
On the general video understanding benchmarks (LVB, Video-MME) that contains complex long video inputs, \model{} significantly outperforms Video-R1 by $3.2$ and $1.6$ with less frame input ($60.5$ and $56.5$) for video inference. 
Our method also outperforms the leading open-source general-purpose MLLM (Qwen-2.5-VL \cite{Qwen2.5-VL}) on LVB with  $92.2\%$ average frames for evaluation. 
This result shows the efficiency and effectiveness of \model{} on reasoning over complex video inputs. 
To sum up, our method achieves the best report performance among all open-source video reasoning methods

\subsection{Analysis}

\begin{figure}
    \centering
    \includegraphics[width=0.75\linewidth]{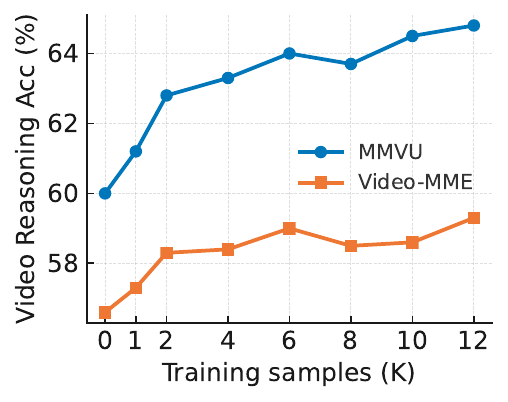}
    \vspace{-0.1in}
    \caption{Analysis on the number of training samples for pure-RL training. }
    \label{fig:training_sample_fig}
\end{figure}

\begin{table}[t]
  \setlength{\tabcolsep}{4pt}
  \centering
  \small        
  \begin{tabular}{l|c|cc}
    \toprule
    \textbf{Reasoning} & \textbf{Data} & \textbf{MMVU} & \textbf{Video-MME} \\
    \midrule
    Zero-shot CoT  & –          & 60.0 & 56.6 \\
    SFT            & 165K       & 63.5 & 55.4 \\ 
    SFT+RL         & 165K+4K  & 63.8 & \textbf{59.3} \\
    \midrule
    \ours{} (Ours) & 6K          & \textbf{64.0} & 59.0 \\
    \bottomrule
  \end{tabular}
  \vspace{-0.05in}
  \caption{Efficiency of our pure-RL training method.}
  \label{tab:rl_comparison}
  \vspace{-0.05in}
\end{table}

\noindent\textbf{Efficiency of the Pure-RL Training.}
In \cref{tab:rl_comparison}, we verify the efficiency and effectiveness of our pure-RL training with zero-shot CoT prompting on base MLLM \cite{Qwen2.5-VL}, large-scale SFT and SFT+RL frameworks \cite{feng2025video}. 
With only 6K training data, our pure-RL method gets on par performance with the large-scale SFT+RL framework (trained on 165k+4K data) with only $3.6\%$ samples, validating the training efficiency of \model{}. 
What's more, our pure RL only requires the ground truth answer as a training signal, which skips the cumbersome data generation process for SFT on the video reasoning task. 
Meanwhile, in \cref{fig:training_sample_fig}, we also analyze the gain in video reasoning performance with different numbers of training samples. 
We observe a sharp performance gain within the first $2K$ samples and continually get improvements on both MMVU and Video-MME until $6K$ samples.
However, the pure-RL training seems to be saturated at $6K$ sample,s and scaling to more samples will degrade or get marginal performance. 
We also show the results with even more training samples in \cref{tab:app_training_data_abla}, which confirms this trend. 
We argue that the base MLLM possesses powerful reasoning capabilities derived from its pretraining and post-training stages. Our pure-RL training method helps the model become familiar with the video QA format and transfers the reasoning ability from language to complex video inputs. 
Thus, we propose saving the excessive training resources and allocating them to the inference stage to improve video reasoning capability while remaining efficient.

\begin{figure*}
    \centering
    \includegraphics[width=1\linewidth]{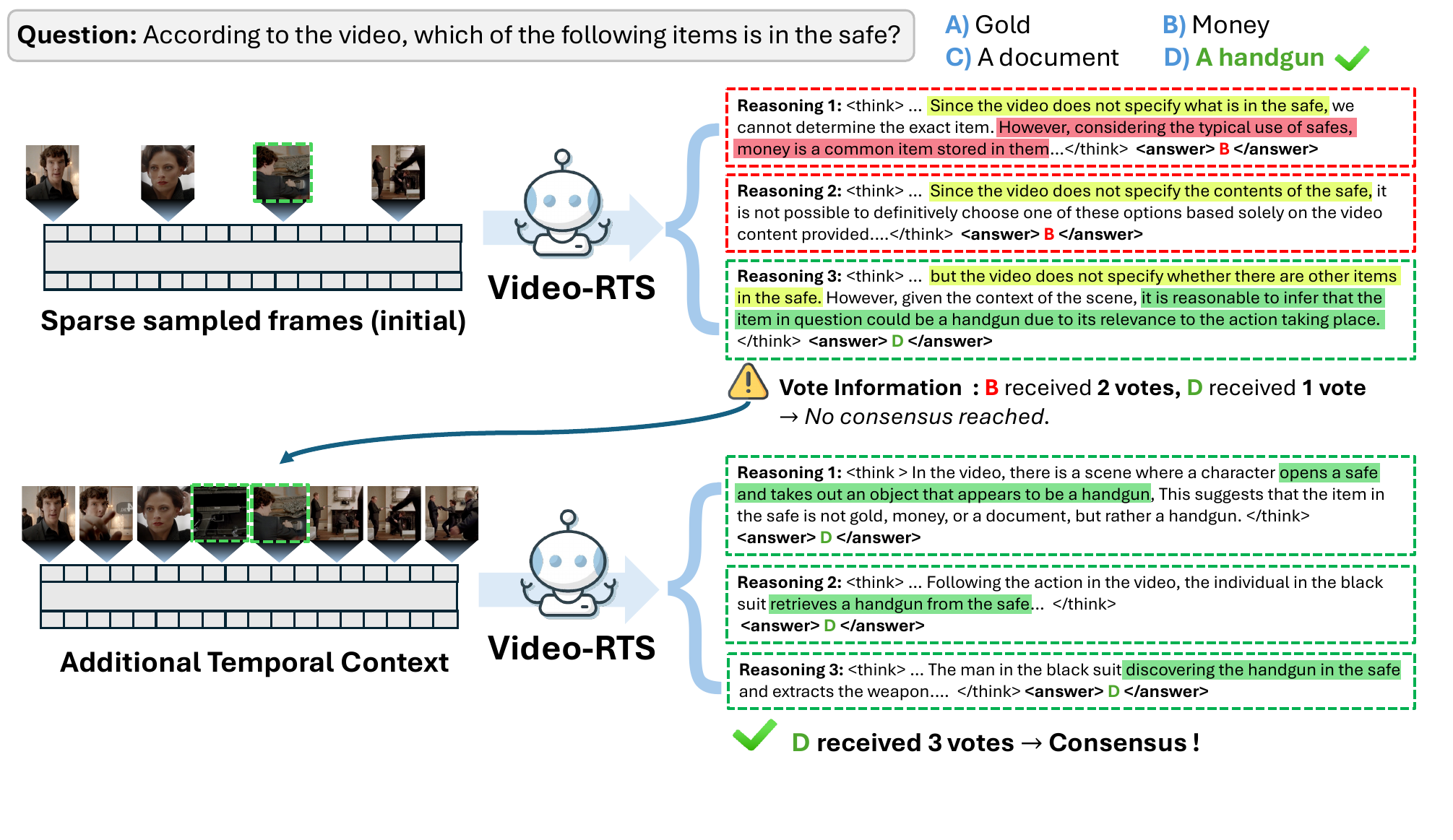}
    \vspace{-0.6in}
    \caption{\textbf{Illustration of dynamic sparse-to-dense reasoning in \ours{}.} \ours{} identifies when the sampled visual information is insufficient for accurately reasoning about the input query (reasoning highlighted in yellow background), often leading to no consensus among intermediate reasoning steps and potentially inaccurate predictions (in red). \ours{} enables the model to adaptively refine its reasoning process (in green), through the proposed dynamic sparse-to-dense reasoning mechanism, achieving accurate and consensus-driven predictions.
    } 
    \label{fig:vis_1}
\end{figure*}

\begin{table}[t]
  \setlength{\tabcolsep}{4pt}
  \centering
  \small        
  \begin{tabular}{l|cc}
    \toprule
    \textbf{Inference Method} & \textbf{MMVU} & \textbf{Video-MME} \\
    \midrule
    Pure-RL (vanilla)          & 64.0 & 59.0 \\
    + Self-Consistency       & 64.6 & 60.6 \\
    + Weighted Voting          & 64.2 & 60.4 \\
    + Self-Reflection        & 64.4 & 59.3 \\
    \midrule
    + S2D Video TTS (Ours)          & \textbf{66.4} & \textbf{63.0} \\
    \bottomrule
  \end{tabular}
  \caption{Comparison of sparse-to-dense video test-time scaling with different test-time scaling strategies. S2D refers to `sparse-to-dense'.}
  \label{tab:testtime_scaling_comparison}
\end{table}

\noindent\textbf{Effectiveness of Video-Specific TTS Design.} 
In \cref{tab:testtime_scaling_comparison}, we compare adaptive video test-time scaling with the popular test-time scaling (TTS) methods: self-consistency \cite{wang2022self}, weighted voting \cite{wan2025reasoningawareselfconsistencyleveraging}, self-reflection~\cite{shinn2023reflexionlanguageagentsverbal}. 
For weighted voting, we use the recent IXC-2.5-Reward \cite{internlmxcomposer2_5_reward} model as the verifier model.
Results show that adaptive video test-time scaling outperforms the existing language-based TTS methods by $2.0\%$ and $2.6\%$ on MMVU and Video-MME benchmarks, respectively. 
This validates the adaptive design of adaptive video test-time scaling improves the reasoning capability for \model{} on complex videos.

\begin{table}[t]
  \setlength{\tabcolsep}{4pt}
  \centering
  \small
  \begin{tabular}{cc|cc}
    \toprule
    \textbf{Pure-RL} & \textbf{S2D Video TTS} & \textbf{MMVU} & \textbf{Video-MME} \\
    \midrule
                &              & 60.0 & 56.6 \\
    \checkmark  &              & 64.0 & 59.0 \\
                & \checkmark   & 63.8 & 59.6 \\
    \checkmark  & \checkmark   & \textbf{66.4} & \textbf{63.0} \\
    \bottomrule
  \end{tabular}
  \caption{Effectiveness of combining our pure-RL and adaptive video test-time scaling design for video reasoning tasks. The baseline model on the top row is zero-shot CoT on base MLLM. }
  \label{tab:ablation_rl_tts}
\end{table}

\noindent\textbf{Vote Count Analysis of Video-TTS.}
In \cref{tab:vote_ablation}, we ablate the effectiveness of vote counts in S2D video TTS design. 
Moving from a single trajectory to five significantly improves the video reasoning performance on MMVU by $2.4\%$ and Video-MME by $4.0\%$. 
This result indicates that five independent samplings already offer sufficient diversity for robust majority agreement while keeping \model{} efficient.
Pushing the pool to $10$ or $20$ adds even more low-probability reasoning chains, making the model's consensus less reliable and performance dips despite higher compute.

\begin{table}[t]
  \setlength{\tabcolsep}{4pt}  
  \centering
  \small
  \begin{tabular}{c|cc}
    \toprule
    \textbf{\#Votes} & \textbf{MMVU} & \textbf{Video-MME} \\
    \midrule
        1 &  64.0  &  59.0 \\
       5  &  66.4  &  63.0 \\
      10  &  66.2  &  63.1 \\
      20  &  65.8  &  62.6 \\
    \bottomrule
  \end{tabular}
  \caption{Ablation on vote count $m$ in S2D Video-TTS. The vote $1$ is \model{} without  test-time scaling. }
  \label{tab:vote_ablation}
\end{table}

\noindent\textbf{RL+TTS Yields Strong Video Reasoning Ability.}
In \cref{tab:ablation_rl_tts}, we showcase the two main components of our method, pure-RL training and sparse-to-dense video test-time scaling, which are complementary for strong video reasoning capability. 
Individually, pure-RL training sharpens the reasoning ability of the base MLLM and brings significant gains compared to the zero-shot CoT prompting baseline. 
adaptive video test-time scaling alone also improves the performance, but its gains are limited by the reasoning capability of the base MLLM. 
When pure-RL training and S2D Video TTS are combined, improvements stack almost additively, pushing accuracy to $66.4\%$ on MMVU and $63.0\%$ on Video-MME.

\noindent\textbf{Qualitative Analysis.}
In \cref{fig:vis_1}, we visualize qualitative results from \model{}. 
Specifically, we show the effectiveness of the sparse-to-dense video inference process of \model{}. 
In this example, given the query \textit{``According to the video, which of the following items is in the safe''}, our model initially attempts to reason using a sparse frame set (i.e., a small number of sampled frames). As shown at the top, the reasoning (highlighted in yellow) lacks sufficient visual evidence, resulting in unclear and inconsistent predictions across multiple runs.
In this case, \model{} dynamically integrates additional frames into the inference process (bottom), allowing for more accurate and consistent reasoning by leveraging concrete visual cues (highlighted keyframes and reasoning steps are marked in green) to answer the visual query.
This visualization showcases the adaptiveness of \model{} during the inference stage and leads to accurate video reasoning.

%% file: sec/6_conclusion.tex
\section{Conclusion}
\label{sec:conclusion}
We introduce \model{}, the first work that systematically explores the combination of reinforcement learning and test-time scaling for video reasoning tasks. 
Instead of using long CoT data for SFT, we utilize a pure-RL training and propose an adaptive video inference strategy that allocates the excessive training compute to test time and improves the video reasoning capability. 
We validate the effectiveness of our model on four popular benchmarks, showing that \model{} outperforms the recent video reasoning model 
by $2.4\%$ in average accuracy with $3.6\%$ training samples and fewer frames for inference.
Importantly, we find that pure-RL training and adaptive video test-time scaling work synergistically, yielding the superior video-reasoning ability of \model{}.

%% file: sec/8_appendix.tex
\appendix

\section*{Appendix}
\label{sec:appendix}

Our appendix consists of Limitations (\cref{sec:limitation}), Additional Implementation Details (\cref{sec:app_imp}), Additional Quantitative Analysis (\cref{sec:app_res}), Detailed Prompts (\cref{sec:prompts}) and License and Artifact Usage (\cref{app:license}).

\section{Limitations}
\label{sec:limitation}
Like other LLM-based video reasoning systems, our method may inherit societal or ethical biases from malicious content in the pretraining data of the base (M)LLMs. 
However, since \model{} mitigates this risk through a consensus-based adaptive inference strategy, where biased outputs under certain sampling conditions can be counteracted by more neutral ones, enhancing fairness. Additionally, our method relies on relatively small-scale video reasoning data, making it more practical to filter harmful samples compared to large-scale SFT and RL setups. In future work, we aim to further investigate data quality and develop fairness-aware video reasoning models.

\section{Additional Implementation Details}
\label{sec:app_imp}
For training data, we sample from the CG-Bench dataset \cite{chen2024cgbenchcluegroundedquestionanswering}, which is a 
All results are under same, fixed random seed with single runs.
For the majority voting in all test-time scaling method, we sample with temperature $ (\tau_i = 0.7 + 0.1\,i)$ and nucleus threshold $(p_i = \max(0.5,\;0.9 - 0.1\,i))$.
Generation is capped at $1024$ tokens and stops early at the sentinel token \texttt{</answer>}.
During evaluation, following Video-R1 \cite{feng2025video}, we set the max frame resolution to $256 * 28 * 28$.

\section{Additional Quantitative Results}
\label{sec:app_res}

\subsection{Training Data Analysis}

We use CG-Bench as the primary training source, which contains $1.2k$ videos with $12k$ multiple-choice question-answering (MCQA) samples. 
Following \citet{yu2025dapoopensourcellmreinforcement, zheng2025deepeyesincentivizingthinkingimages}, we filter out the samples that are either too easy or too difficult to support more effective learning.
Specifically, we generate 8 responses per sample and calculate their difficulty based on accuracy; samples with accuracy of 0 or 1 are excluded. 
To ensure a diverse set of video types, we then randomly sample an equal number of examples from each video, resulting in a final training set of $6K$ samples.
In \cref{tab:app_training_data_abla}, we report the impact of different training data selection strategies and show that the manually annotated CG-Bench data \cite{chen2024cgbenchcluegroundedquestionanswering} outperforms the larger LLaVA-Video-178K data \cite{zhang2024videoinstructiontuningsynthetic}. 
Meanwhile, filtering out overly easy or hard samples benefits RL training, further improves performance, and speeds up the training process. 

\begin{table}[h]
  \setlength{\tabcolsep}{3pt}
  \centering
  \small        
  \begin{tabular}{lcc|cc}
    \toprule
    \textbf{Data Source} & \textbf{\# Sample}  & \textbf{Filter} & \textbf{MMVU} & \textbf{V-MME} \\
    \midrule
    Zero-shot  & –        &     & 60.0 & 56.6 \\
    LLaVA-V            & 35k      &    & 63.5 & 55.4 \\ 
    CG-Bench         & 12k   &   & 63.2 & \textbf{58.3} \\
    LLaVA-V          & 6k   & \checkmark  & 63.8 & \textbf{59.3} \\
    CG-Bench         & 6k   & \checkmark  & 64.0 & \textbf{59.0} \\
    \bottomrule
  \end{tabular}
  \vspace{-0.05in}
  \caption{Training data analysis.}
  \label{tab:app_training_data_abla}
  \vspace{-0.05in}
\end{table}

\subsection{Evidence for Performance Saturation with Increasing Training Samples}
\label{app:saturation}

\cref{fig:training_sample_fig} already suggests a saturation point after roughly $6K$ training examples.
To further investigate this trend, we extend training to $12K$ samples by combining the original \textsc{CG-Bench} with carefully deduplicated examples from the public \textsc{LLaVA-Video-178K} corpus \citep{zhang2024videoinstructiontuningsynthetic}.
\cref{tab:train_scale} shows that
beyond 6 K examples, accuracy gains plateau (and occasionally regress), highlighting the importance of data-efficient methods such as ours.

\begin{table}[h]
    \centering
    \label{tab:train_scale}
    \begin{tabular}{lccc}
        \toprule
        \# Training Sample              & 15 K & 20 K & 25 K \\
        \midrule
        MMVU          & 64.1 & 63.8 & 63.5 \\
        Video-MME     & 58.8 & 59.1 & 59.2 \\
        \bottomrule
    \end{tabular}
        \caption{RL training with more samples.}
\end{table}

\subsection{Random Seeds Analysis}
To verify that our results are not sensitive to the particular subset chosen (random sample $6K$ samples from CG-Bench), we repeat the sampling procedure with five distinct random seeds.
After the RL fine-tuning step, the MMVU accuracy averages
\( 64.2\,\pm\,0.3 \) across the five runs, closely matching the
64.0 reported in \cref{tab:ablation_rl_tts}.
This confirms that the observed performance is robust to the exact choice of training samples.

\begin{table*}[t!]
\centering
\begin{minipage}{2\columnwidth}\vspace{20mm}    
    \centering
    \caption{\textbf{\model{} detailed prompt.}}
    \begin{tcolorbox} 
        \centering
        \hspace{-6mm}
        \begin{tabular}{p{0.99\columnwidth}}
        \begin{minipage}{0.99\columnwidth}
        \textbf{User} \\
        A conversation between User and Assistant. The user asks a question, and the Assistant solves it. The assistant first thinks about the reasoning process in the mind and then provides the user with the answer. The reasoning process and answer are enclosed within <think> </think> and <answer> </answer> tags, respectively, i.e., <think> reasoning process here </think><answer> answer here </answer>. \\ 
        Video: \textcolor{blue}{\texttt{Video Tokens}} \\ 
        Question: \textcolor{blue}{\texttt{Question}} \\
        Options: 
        A: \textcolor{blue}{\texttt{Option-A}}. B: \textcolor{blue}{\texttt{Option-B}}. C: \textcolor{blue}{\texttt{Option-C}}. D: \textcolor{blue}{\texttt{Option-D}}..... \\
        Please provide only the single option letter (e.g., A, B, C, D, etc.) within the <answer> </answer> tags.')\\
        \rule[0.25\baselineskip]{\textwidth}{1pt}
        \end{minipage}
        \end{tabular}
    \end{tcolorbox}
\label{tab:prompt}
\end{minipage}
\end{table*}

\subsection{Sparse-to-Dense Video TTS on Other MLLM Backbone}

To further validate the effectiveness of our framework, we conduct additional experiments using InternVL-3\cite{zhu2025internvl3exploringadvancedtraining}, which is also a leading MLLM that has already been trained on reasoning data using RL. We directly evaluate our S2D Video TTS method using the InternVL-3-8B model on a video reasoning benchmark and demonstrate that our S2D Video TTS can achieve a 1.5\% improvement in MMVU accuracy, validating the generalization ability of Video-RTS.

\section{Detailed Prompts}
\label{sec:prompts}
We provide detailed prompts for \model{} in \cref{tab:prompt}. We use the same format of prompts for training and evaluation.

\section{License and Artifact Usage}
\label{app:license}

\subsection{License}
We use standard licenses from the community and provide the following links to the licenses for the datasets, codes, and models that we used in this paper:

\vspace{3pt}
\noindent\textbf{TRL:} \href{https://github.com/huggingface/trl/blob/main/LICENSE}{Apache}

\vspace{3pt}
\noindent\textbf{Video-R1:} \href{https://github.com/tulerfeng/Video-R1}{Video-R1}

\vspace{3pt}
\noindent\textbf{Open-R1-Video:} \href{https://github.com/Wang-Xiaodong1899/Open-R1-Video/blob/main/LICENSE}{Apache}

\vspace{3pt}
\noindent\textbf{Qwen-2.5-VL:} \href{https://github.com/QwenLM/Qwen2.5-VL/blob/main/LICENSEe}{Apache}

\vspace{3pt}
\noindent\textbf{CG-Bench:} \href{https://github.com/CG-Bench/CG-Bench}{CG-Bench}

\vspace{3pt}
\noindent\textbf{Video-MMMU:} \href{https://github.com/EvolvingLMMs-Lab/VideoMMMU/blob/main/LICENSE}{MIT}

\vspace{3pt}
\noindent\textbf{Video-MME:} \href{https://github.com/MME-Benchmarks/Video-MME}{Video-MME}

\vspace{3pt}
\noindent\textbf{MMVU:} \href{https://github.com/yale-nlp/MMVU.git}{MMVU}

\vspace{3pt}
\noindent\textbf{LongVideoBench:} \href{https://github.com/longvideobench/LongVideoBench}{CC-BY-NC-SA 4.0}

\subsection{Artifact Usage}
The use of existing artifacts is consistent with their intended use in We will make our code and models publicly accessible and all created artifacts will be only for research purposes and should not be used outside of research contexts.